\documentclass[letterpaper]{article} % DO NOT CHANGE THIS
\usepackage{aaai25}  % DO NOT CHANGE THIS
\usepackage{times}  % DO NOT CHANGE THIS
\usepackage{helvet}  % DO NOT CHANGE THIS
\usepackage{courier}  % DO NOT CHANGE THIS
\usepackage[hyphens]{url}  % DO NOT CHANGE THIS
\usepackage{graphicx} % DO NOT CHANGE THIS
\usepackage{natbib}  % DO NOT CHANGE THIS AND DO NOT ADD ANY OPTIONS TO IT
\usepackage{caption} % DO NOT CHANGE THIS AND DO NOT ADD ANY OPTIONS TO IT
\usepackage{algorithm}
\usepackage{algorithmic}

\usepackage{newfloat}
\usepackage{listings}
\DeclareCaptionStyle{ruled}{labelfont=normalfont,labelsep=colon,strut=off} % DO NOT CHANGE THIS
\floatstyle{ruled}
\newfloat{listing}{tb}{lst}{}
\floatname{listing}{Listing}
\usepackage{booktabs}
\usepackage{multirow}
\usepackage{adjustbox}
\usepackage{xcolor} 

\title{Evaluating LLM Reasoning in the Operations Research Domain with ORQA}
\author {
    Mahdi Mostajabdaveh \textsuperscript{\rm 1},
     Timothy T. Yu \textsuperscript{\rm 1},
     Samarendra Chandan Bindu Dash \textsuperscript{\rm 1,2},
     Rindranirina Ramamonjison \textsuperscript{\rm 1},
     Jabo Serge Byusa \textsuperscript{\rm 1},
     Giuseppe Carenini \textsuperscript{\rm 3},
     Zirui Zhou \textsuperscript{\rm 1},
     Yong Zhang \textsuperscript{\rm 1},
}
\affiliations {
    \textsuperscript{\rm 1}Huawei Technologies Canada, 4321 Still Creek Dr, Burnaby, BC, V5C 6S7, Canada  \\
    \textsuperscript{\rm 2} University of Toronto, 40 George St, Toronto, ON, M5S 2E4, Canada \\
    \textsuperscript{\rm 3}University of British Columbia,
    2366 Main Mall, Vancouver, BC, V6T 1Z4, Canada\\ 
    \{mahdi.mostajabdaveh1, timothy.yu, rindranirina.ramamonjison, jabo.byusa, zirui.zhou, yong.zhang3\}@huawei.com, carenini@cs.ubc.ca, dashsam1@cs.toronto.edu.
}

\begin{document}

\maketitle

\begin{abstract}
In this paper, we introduce and apply \textbf{O}perations \textbf{R}esearch \textbf{Q}uestion \textbf{A}nswering (ORQA), a new benchmark, to assess the generalization capabilities of Large Language Models (LLMs) in the specialized technical domain of Operations Research (OR). This benchmark is designed to evaluate whether LLMs can emulate the knowledge and reasoning skills of OR experts when given diverse and complex optimization problems. The dataset, crafted by OR experts, presents real-world optimization problems that require multi-step reasoning to build their mathematical models. Our evaluations of various open-source LLMs, such as LLaMA 3.1, DeepSeek, and Mixtral reveal their modest performance, indicating a gap in their aptitude to generalize to specialized technical domains. This work contributes to the ongoing discourse on LLMs' generalization capabilities, providing insights for future research in this area. The dataset and evaluation code are publicly available\footnote{\url{https://github.com/nl4opt/ORQA}}.
\end{abstract}

% Uncomment the following to link to your code, datasets, an extended version or similar.
%
% \begin{links}
%     \link{Code}{https://aaai.org/example/code}
%     \link{Datasets}{https://aaai.org/example/datasets}
%     \link{Extended version}{https://aaai.org/example/extended-version}
% \end{links}

\section{Introduction}
The ability of Large Language Models (LLMs) to follow human instructions and perform diverse tasks has made them an exciting area of investigation. Moreover, the considerable interest in adopting LLMs across various complex technical domains (e.g., medicine \citep{yunfan2024, zhou2023survey}) highlights their potential for significant societal impact. A particularly compelling driver for this adoption is the potential of LLMs to automate many tasks, reducing human intervention and improving productivity. However, as LLMs are becoming integrated into the workflow of various industries, it is important to thoroughly understand their capabilities and limitations \citep{aisha2024, jawid2023, truong2023language}. Of particular interest is their ability to reason and perform new challenging tasks across different domains, which are reported critical limitations of LLMs \citep{konstantine2023, generalizationLimitations2023}. Our work addresses this need by introducing a new benchmark dataset and applying it to assess these limitations. % of LLMs.

To evaluate an LLM's ability to generalize to a new domain, we focus on the field of Operations Research (OR). The choice of this domain is deliberate and significant. First, OR is important for making decisions in various industries \citep{fotios2023}. Second, there are many types of optimization problems applied to real-world applications ranging from production scheduling \citep{mostajabdaveh2022two} to creating efficient delivery routes for trucks \citep{vidal2020concise}. Third, optimization modeling presents a unique challenge due to the expert-level knowledge and reasoning skills it requires \citep{hillier2015}, adding a layer of complexity to the automation of this task. Some recent studies also report modest performances of SOTA LLMs such as GPT-4 and Llama2 for optimization model building tasks \citep{OptiMUS, multiAgentModeling2024}. 
Finally, OR is a niche field with limited publicly available text corpora or optimization model code \citep{Chain-of-Experts, OptiMUS}, making it an ideal testbed to assess the generalizability of LLMs, reducing the risk of data contamination.

\begin{figure*}[h!]
  \centering
  \includegraphics[width=0.8\linewidth]{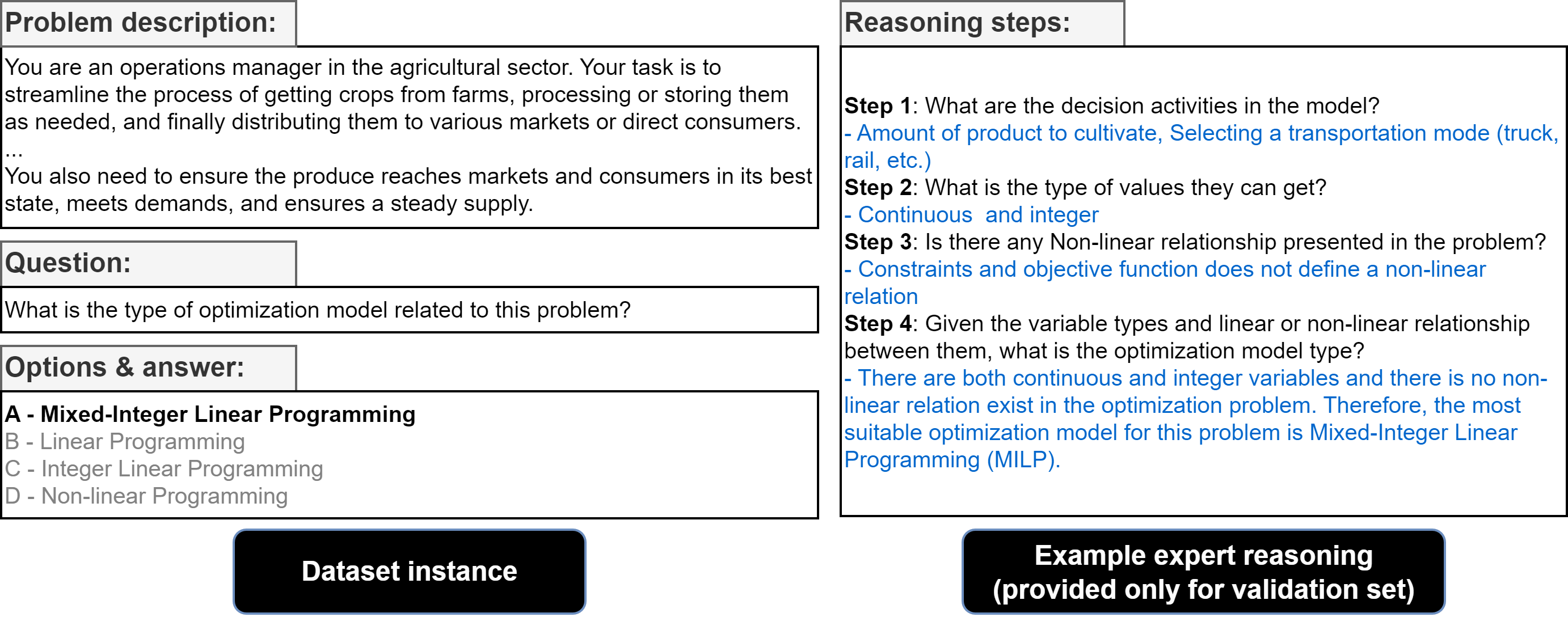}
\caption{\textbf{Left}: Dataset instance containing the description, question, options, and answer (bold). \textbf{Right}: Example reasoning steps needed to answer the simple question.}
\label{Fig:example}
\end{figure*}

To assess an LLM's knowledge and reasoning skills on unseen, diverse, and complex optimization problems, we propose ORQA (\textbf{O}perations \textbf{R}esearch \textbf{Q}uestion \textbf{A}nswering), a new multi-choice Question Answering (QA) benchmark dataset, crafted and verified by OR experts. % The dataset spans a variety of application ``sub-domains", including marketing and logistics, among others. COMMENT: we repeat this point too many times
Each dataset instance (Figure~\ref{Fig:example}; left) presents a natural language description of an optimization problem along with a question that requires multi-step reasoning (Figure~\ref{Fig:example}; right) to answer correctly.

Our study is significant for several reasons. First, it contributes to the ongoing dialogue on the generalizability of LLMs. Although many benchmarks claim that LLMs can replicate expert-level knowledge across various technical domains, the actual extent of these models' generalization capabilities remains an open question \citep{norah2024}. Our benchmark offers a new perspective by focusing on a specialized technical domain that lacks a large-scale, high-quality dataset. To the best of our knowledge, this is the first multi-choice QA dataset in the field of Operations Research.
Second, this study has implications for understanding the potential and limitations of LLMs in automating tasks in niche technical fields like OR, where expert knowledge and complex multi-step reasoning are crucial. Optimization modeling requires access to specialized OR experts, which is often impractical for most potential users due to the associated costs.

We choose to focus on translating textual problem descriptions into mathematical optimization models rather than directly solving optimization problems. Even specialized AI models struggle with scalability and generalization when solving simple OR problems \citep{joshi2022learning}. We also highlight the challenges of constructing a technical dataset like ORQA. Samples of optimization problems are inherently complex and require significant time and effort to create and verify. Moreover, ensuring correctness demands annotators with graduate-level education or extensive experience in OR modeling, making the process expensive, time-consuming, and labor-intensive.

\section{Related Work}
\paragraph{LLMs applied to operations research:} 
Within the field of OR, LLMs are being investigated for their ability to formulate optimization models \citep{fan2024artificial}. \citet{ramamonjison2022augmenting} proposed using generative models to automate the formulation of OR problems from natural language. Building on this, \citet{ramamonjison2023nl4opt} introduced methods for recognizing entities and parsing optimization formulations from text. However, these works primarily utilize a toy dataset of elementary linear programming word problems. More recently, \citet{OptiMUS} proposed the NLP4LP dataset, which includes 332 instances with structured natural language descriptions (SNOP), parameter data values, optimal value, and optimal solution. However, the majority of the problems are still toys, their description is technical, with mathematical notation, and lacks context from real applications. \citet{Chain-of-Experts} introduced the ComplexOR dataset with only 37 optimization problems. This dataset includes NL descriptions, optimization models, and several input data. While their problem descriptions are context-aware, they often mention the related optimization problem by name (e.g., lot-sizing problem with setup), and only cover a narrow range of application domains. With more than 1.5k instances, our dataset is the largest, offering significant value through its depth, rigor, realism, and diversity.

Another disadvantage of existing mentioned datasets is they require the optimization model to be solved for their evaluation. LLMs must generate a model code from the NL description, which is then fed to a solver with data to obtain the optimal value or solution. The evaluation focuses solely on the correctness of the optimal value or solution, presenting two key limitations: (1) it is an end-to-end evaluation that cannot differentiate between minor notation errors and entirely incorrect structures; (2) it cannot distinguish between errors in code generation and errors in model formulation. \cite{OptiMUS} show that coding errors account between 21\% to 31\%. 
In contrast, ORQA tackles more complex optimization tasks across diverse application scenarios. Our problem descriptions are context-aware, relevant to real applications, and free of OR jargon and mathematical notations. Additionally, ORQA is  multiple-choice question answering dataset offers a straightforward evaluation that is independent of model code and does not require solvers.

\paragraph{Multi-choice question answering:} 
The multi-choice QA task involves receiving a question along with several candidate options and selecting the correct answer. Many such datasets have been made publicly available \citep{talmor2019, mihaylov2018, clark2020}. However, the complexity of current reasoning benchmarks has been called into question \citep{khatun2024, valmeekam2023, sawada2023}, motivating the creation of more challenging benchmarks that surpass basic commonsense reasoning \citep{kweon2024, sawada2023}. 
While multi-choice QA is a well-studied NLP task, ORQA stands out as a handcrafted, expert-curated dataset in the technical field of OR. This domain is notably underrepresented in current benchmarks and demands deep optimization knowledge. Moreover, tasks in ORQA require identifying optimization model components and their interrelationships, which necessitates multi-step reasoning.

\begin{figure*}[h!]
  \centering
  \includegraphics[width=\linewidth]{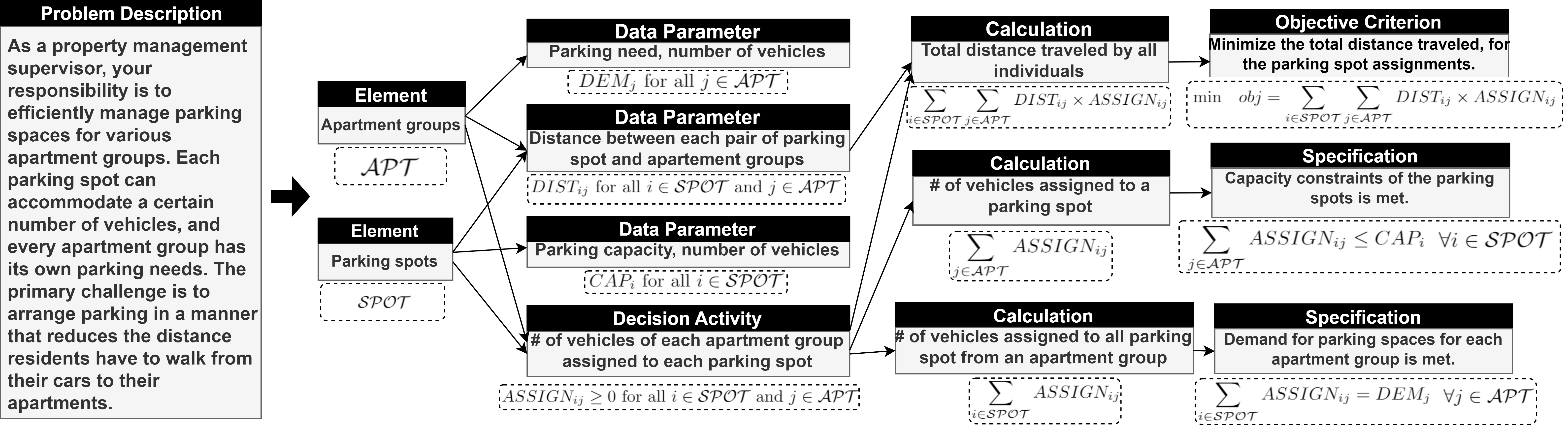}
\caption{An example of optimization problem components, their relationships, and corresponding mathematical formulations.}
\label{fig:components_exp}
\end{figure*}

\paragraph{LLM reasoning capabilities and limitations:} 
\citet{zhou2024self} demonstrated that LLMs can be prompted to reason, leading to improved performance and insights into how decisions are made. Techniques for this purpose include multi-step chained prompts \citep{yoran2023answering, kojima2022large}, single-step chain-of-thought \citep{kojima2022large}, tree-of-thought \citep{yao2024tree}, and chain-of-thought with self-consistency and verification \citep{zhao2023verify}.
While these methods are promising, limitations persist in leveraging and evaluating LLMs' reasoning capabilities. Evaluating the faithfulness of reasoning, as highlighted by \citet{lanham2023measuring}, is a significant challenge. Most existing reasoning benchmarks are overly simplistic, indicating a need for more complex benchmarks \citep{Valmeekam22, philippe2023}. Given this need, our ORQA benchmark dataset is designed to test the reasoning abilities of LLMs in the demanding OR context.

\section{Task: Identifying Optimization Model Characteristics}
\subsection{Task background and motivation}
An optimization model is a mathematical representation of a decision-making problem. 
%It is often formulated into code using a modeling language which is then input to an optimization solver (e.g., Gurobi \citep{gurobi2023}) to obtain the optimal solution.
Optimization models are constructed using various components. These components include elements, decision activities, data attributes, calculations, objective criteria, and specifications \citep{modelling}. The first and most crucial step in formulating an optimization model is to identify the components of the model and understand their relationships, as any error in this step will result in an incorrect model. ORQA focuses on identifying these components and their relations from the natural language description of the optimization problem by asking questions such as ``What are the decision activities of the optimization problem?" and ``Which data parameters are participating in the objective criterion?".

To illustrate these components and their relationships, we refer to the parking spot assignment example problem described in Figure~\ref{fig:components_exp}. The figure demonstrates how the extraction of optimization problem components and their relationships can be directly used to formulate the optimization problem mathematically. Unique apartment units and parking spots are the elements of this example problem and their data attributes directly map to sets (i.e., apartment groups, parking spots) and data parameters (i.e., parking need, number of vehicles, etc.). The decision activities are direct actions in the system and define the optimization variables (i.e., assign vehicles to parking spots). These three model components are combined to form the utility/cost function to be maximized/minimized (i.e., minimize total distance). They also form the specifications that define business rules or system limitations, which lead to optimization constraints.

\subsection{Task definition}
We propose a multi-choice QA task to identify the components of the optimization problem, their attributes, and relationships, from a given natural language problem description. This is a highly complex task requiring multi-step reasoning as these components have multiple layers of interaction and dependencies \citep{hillier2015}. For example, identifying the objective criteria involves not only recognizing the objective measure and sense, but also determining the specific data attributes and decision activities that influence it. Figure~\ref{Fig:components} in Appendix~\ref{app:components} illustrates these complex relationships between problem components.
%Similarly, identifying specifications entails understanding the constraints, the elements they are applied to, and how they interact with the decision activities, parameters and calculations. 

\subsection{Task characteristics}
The task we propose is complex not only due to the heavily mathematical nature of the field of OR, but also the complexity of the optimization models the dataset is built upon. The complexity is directly related to the number of components in the corresponding mathematical model. We describe in Section 4.2 our approach to ensure a standard level of complexity during our dataset creation process.

Additionally, the task is difficult due to the under-representation of open-sourced OR data during LLM training. The findings from \citet{kandpal2023large} and \citet{alex2023} align with our claim that scarcity in optimization modeling-related data would make this task challenging for LLMs. In this regard, we consider optimization modeling as a task that requires long-tail knowledge.

\section{ORQA Dataset}
\subsection{Dataset overview}
Each dataset instance contains the following (see Figure~\ref{Fig:example}, left):

\begin{enumerate}
  \item A \emph{CONTEXT} describing an optimization problem as a case study using natural language,
  \item A \emph{QUESTION} asking about the problem specifications (e.g., objective criterion or constraints), the underlying components of the model (e.g., the elements participating in the optimization), or the structure and logic of the optimization model (e.g., the logical relation between two components),
  \item A list of \emph{OPTIONS} for the answer, which was created by OR experts to make the question challenging. The LLM must select the correct answer from a list of four options.
  \item The correct \emph{TARGET ANSWER}.
\end{enumerate}

Table~\ref{tab:dataset_stat} presents the characteristics of the dataset. A wide range of application domains are represented within ORQA ranging from common problems (e.g., Traveling Salesman Problem) to niche problems (e.g., multi-period production planning of a drone manufacturing company). ORQA is comprised of a total of 20 application domains each represented by at least three problems and 60 to 90 multi-choice questions. Some of these domains include healthcare, urban design, human resources, petroleum, and sales.

\begin{table}[]
\centering
\begin{tabular}{l|c}
\hline
Characteristics & ORQA \\
\hline
Number of instances & 1513 \\
Test/validation split & $1468 / 45$ \\
Average input length (words) & 231 \\
Number of domains & 20 \\
\hline
\end{tabular}
\caption{ORQA dataset statistics.}
\label{tab:dataset_stat}
\end{table}

ORQA is comprised of 1513 data instances with 45 instances allocated as the validation set. The validation set provides the in-context learning (ICL) examples used for few-shot prompting. We include expert-written reasoning steps for the instances in the validation set.

\paragraph{Question type.} The questions in this QA task can be split into 11 question types that have been derived from three critical skills of optimization modeling, as described in detail in Table~\ref{tab:Q_category} (Appendix~\ref{app:categories}). Specifically, (A) understanding the high-level problem specifications, (B) identifying entities of the corresponding optimization model, and (C) identifying relationships between components. For examples and additional details of the 11 question types, please refer to Table~\ref{tab:Q_category} in Appendix~\ref{app:categories}.

\subsection{Dataset creation}
% The decision-making problems within our dataset along with their corresponding question and answers have been carefully curated and verified by a team of five OR experts. 
The dataset was carefully created and verified by a team of five experts with extensive experience in optimization modeling: a Bachelor's graduate, two Master's graduates, and two PhDs.
An example of a data instance is shown in Figure~\ref{Fig:example}. The experts went through a joint training session that provided them with step-by-step instructions created by the lead OR expert for the dataset creation process. In the training session, all experts annotated the same three data instances to ensure that they were aligned. 
Then, each OR expert was assigned a subset of problems to create independently. The OR experts used available optimization models from OR textbooks, academic journals, and online code repositories as the initial source and they substantially modified and diversified them in a multi-step, human-led dataset creation process.
Due to the labor-intensive process of creating this dataset, we did not have OR experts overlap in creating the same data instances. Instead, we prioritized developing a larger benchmark and relied on our training session and a rigorous verification stage (as described below) to ensure consistency in difficulty and quality.

\begin{figure}[h!]
  \centering
  \includegraphics[width=\linewidth]{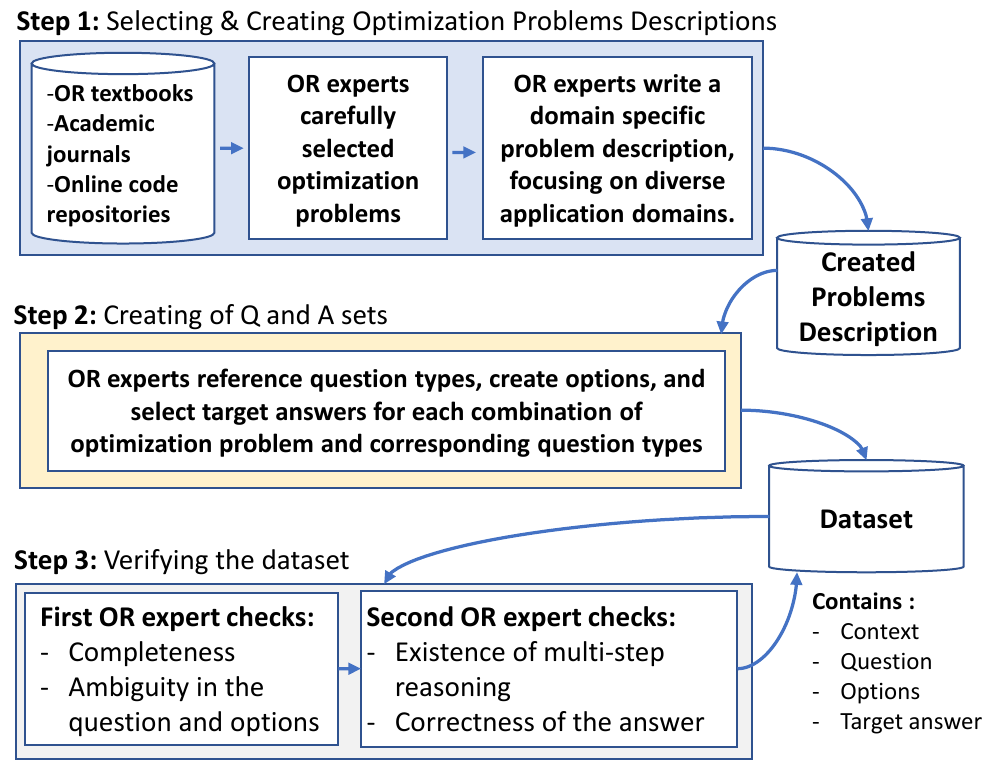}
\caption{Selection, creation, and verification process for the ORQA dataset.}
\label{fig:datasetCreation}
\end{figure}

The selection, creation, and verification process was comprised of three steps and is detailed in Figure~\ref{fig:datasetCreation}. In step 1, problems were filtered based on a set of criteria, which was defined prior to the selection process. The criteria for selection include the problem complexities, diversity in the resulting models types, and practicality for real-world use cases. To regulate the complexity of the dataset, OR experts were instructed to select problems where the number of components in their mathematical model is within pre-defined limits (e.g., the number of decision variables should be between 2 and 7). As a result, the average and standard deviation of the model components are as follows: sets $1.97/0.89$, parameters $4.08 / 2.19$, variables $3.14 / 2.29$, objectives $1 / 0.0$, and constraints $4.60/3.21$.

To regulate problem diversity, each OR expert was given a set of application domains and instructed to select at least three problems from each domain. 
%ORQA showcases a broad range of application domains, ranging from common problems (e.g., variants of the traveling salesman problem) to more niche problems (e.g., multi-period production planning for cement). 
% As described in Table~\ref{tab:dataset_stat}, ORQA comprises a total of 20 application domains. 
Given that most optimization models lack concrete natural language descriptions, our OR experts crafted handwritten problem descriptions and, where applicable, modified the problem context to ensure coverage across diverse application domains.
%Thereby, OR experts created variants of the same mathematical model in different domains. 
Once the OR expert was satisfied with the problem description, they would verify that the natural language description exactly mapped to the original mathematical model.

In step 2 (Figure~\ref{fig:datasetCreation}), two OR experts were responsible for creating and annotating the question, target options, and target answer of each instance of data. Questions were designed to promote multi-step reasoning and they were created along with the options and target answer by referencing both the mathematical model and problem description. For cases where multiple modeling approaches are possible, OR experts ensured that incorrect options were truly incorrect considering all different models.

Finally, in step 3 (Figure~\ref{fig:datasetCreation}), each instance was verified by another cohort of two OR experts. These experts verified that each data instance was complete and correct, devoid of ambiguities in its question and options, required multi-step reasoning, and free of sensitive information (e.g., real-world people and organizations names).% Throughout the data creation and verification process, the experts carefully avoided and screened for any mention of sensitive information (e.g., real-world people and organizations).

\setlength{\tabcolsep}{1mm}
\begin{table}[]
\small
\centering
\begin{tabular}{llllll}
\hline
\multicolumn{1}{c}{\multirow{2}{*}{Model}} & \multicolumn{3}{c}{Standard (Acc)} & \multicolumn{2}{c}{CoT (Acc)} \\ \cline{2-6} 
\multicolumn{1}{c}{} & \multicolumn{1}{c}{\textbf{\begin{tabular}[c]{@{}c@{}}0-\\ shot\end{tabular}}} & \multicolumn{1}{c}{\textbf{\begin{tabular}[c]{@{}c@{}}1-\\ shot\end{tabular}}} & \multicolumn{1}{c|}{\textbf{\begin{tabular}[c]{@{}c@{}}3-\\ shot\end{tabular}}} & \multicolumn{1}{c}{\textbf{\begin{tabular}[c]{@{}c@{}}0-\\ shot\end{tabular}}} & \multicolumn{1}{c}{\textbf{\begin{tabular}[c]{@{}c@{}}1-\\ shot\end{tabular}}} \\ \hline
\multicolumn{1}{l|}{Llama3.1-8B-I} & 0.588 & 0.615 & \multicolumn{1}{l|}{0.618} & 0.563 & 0.324 \\
\multicolumn{1}{l|}{Llama3.1-70B-I} & 0.702 & 0.721 & \multicolumn{1}{l|}{0.735} & 0.689 & 0.292 \\
\multicolumn{1}{l|}{Llama3.1-405B-I} & 0.723 & 0.753 & \multicolumn{1}{l|}{0.772} & 0.695 & 0.360 \\
\multicolumn{1}{l|}{Llama3-8B-I} & 0.535 & 0.573 & \multicolumn{1}{l|}{0.592} & 0.530 & 0.364 \\
\multicolumn{1}{l|}{Llama3-70B-I} & 0.676 & 0.716 & \multicolumn{1}{l|}{0.710} & 0.671 & 0.448 \\
\multicolumn{1}{l|}{Llama2-7B-Chat} & 0.368 & 0.375 & \multicolumn{1}{l|}{0.403} & 0.368 & 0.282 \\
\multicolumn{1}{l|}{Llama2-13B-Chat} & 0.409 & 0.437 & \multicolumn{1}{l|}{0.454} & 0.432 & 0.313 \\
\multicolumn{1}{l|}{Llama2-70B-Chat} & 0.526 & 0.552 & \multicolumn{1}{l|}{0.589} & 0.518 & 0.372 \\
\multicolumn{1}{l|}{FLAN-T5-XXL-11B} & 0.503 & - & \multicolumn{1}{l|}{-} & 0.457 & - \\
\multicolumn{1}{l|}{Falcon-7B-I} & 0.245 & 0.246 & \multicolumn{1}{l|}{0.245} & 0.242 & 0.243 \\
\multicolumn{1}{l|}{DeepSeek-M-7B-I} & 0.478 & 0.552 & \multicolumn{1}{l|}{0.559} & 0.379  & 0.514 \\
\multicolumn{1}{l|}{NuminaMath-7B} & 0.250 & 0.484 & \multicolumn{1}{l|}{0.525} & - & 0.290 \\
\multicolumn{1}{l|}{Mistral-7B-I-v0.1} & 0.467 & 0.475 & \multicolumn{1}{l|}{0.483} & 0.460 & 0.407 \\
\multicolumn{1}{l|}{Mistral-7B-I-v0.3} & 0.523 & 0.555 & \multicolumn{1}{l|}{0.555} & 0.539 & 0.543 \\
\multicolumn{1}{l|}{Mixtral-8x7B-I-v0.1} & 0.588 & 0.606 & \multicolumn{1}{l|}{0.612} & 0.565 & 0.565 \\ \hline
\end{tabular}
\caption{Accuracy for each model across different prompting strategies. In model names, \textit{I} stands for Instruct. Empty entries for FLAN-T5 are due to engineering limitations as a result of prompts exceeding the LLM’s input token limit. Due to the slow generation speed of the NuminaMath model, which took around 10 days to generate reasoning steps, we skipped the 0-shot CoT experiment. }
\label{tab:main}
\end{table}

\section{Experiment Setup for Evaluation}
To provide an initial assessment of the difficulty of ORQA and gain insights, we ran a series of experiments to benchmark different LLM models using various prompting strategies.

\paragraph{Baseline models.} Specifically, we run inference on open-source LLMs such as the 11B parameter FLAN-T5 XXL \cite{chung2024scaling}, Falcon-7B-Instruct \citep{falcon40b}, Mistral model series (Jiang et al., 2023), Mixtral series \citep{jiang2024mixtral}, as well as Llama2, 3 and 3.1 models of different sizes \citep{llama2, llama3s}. We made a deliberate decision not to use closed-source LLMs in the spirit of scientific reproducibility. Because when prompting an LLM through APIs, there is unquantifiable uncertainty in how inputs and outputs are processed. Moreover, the LLMs may change or be deprecated, potentially invalidating our results. Model endpoints are commonly deprecated as newer models are made available \footnote{https://platform.openai.com/docs/deprecations}.
%For instance, Text-Davinci-003 and GPT-3.5-Turbo-0301 are deprecated\f{https://platform.openai.com/docs/deprecations}.

We evaluated each model on the 1468 instances of data from the test set for its standard and CoT prompting capabilities in both zero-shot and few-shot settings, as described in Section 3. Table~\ref{tab:main} outlines the accuracy performance of the LLMs evaluated using this benchmark. Furthermore, we report the average F1 scores in Table~\ref{tab:F1} in Appendix~\ref{app:F1}. As a preliminary human baseline, a single expert with a related Ph.D. achieved 93\% accuracy on a random set of 100 instances without any in-context learning examples.

\paragraph{Prompting strategies.} We evaluate the LLM's ability to reason with different prompting strategies. First, all-at-once (standard) prompting evaluates the LLM's robustness to perform these reasoning steps without explicitly prompting it to reason. Conversely, CoT prompting is implemented as a two-step approach following similar works on multi-choice QA \citep{COT, yoran2023answering, kojima2022large}. Specifically, the first prompt elicits the LLM to reason ``step-by-step". Then, the generated reasoning is added to the prompt and given to the LLM to generate the final answer. Both standard prompting and CoT are evaluated for zero-shot and few-shot capabilities, where few-shot prompts are created by randomly sampling instances with the same question type from the validation split. Note that for ICL examples, ground-truth reasonings are excluded in standard prompting but included in CoT.

\paragraph{Evaluation protocol.} Our custom evaluation framework takes inspiration from \citet{robinson2023leveraging} and binds each option to a symbol (i.e., A, B, C, D). We further discovered that appending the prompt ``Therefore, among A through D, the answer is (" to the end reliably guided all tested LLMs to output the required format. We bind each answer to a symbol to avoid punishing models that display the proper reasoning ability to reach the correct solution, but hallucinate or cannot output one of the options exactly. We calculate the accuracy by comparing the exact match between the generated answer and the corresponding ground truth.

\begin{figure}[h!]
  \centering
  \includegraphics[width=\linewidth]{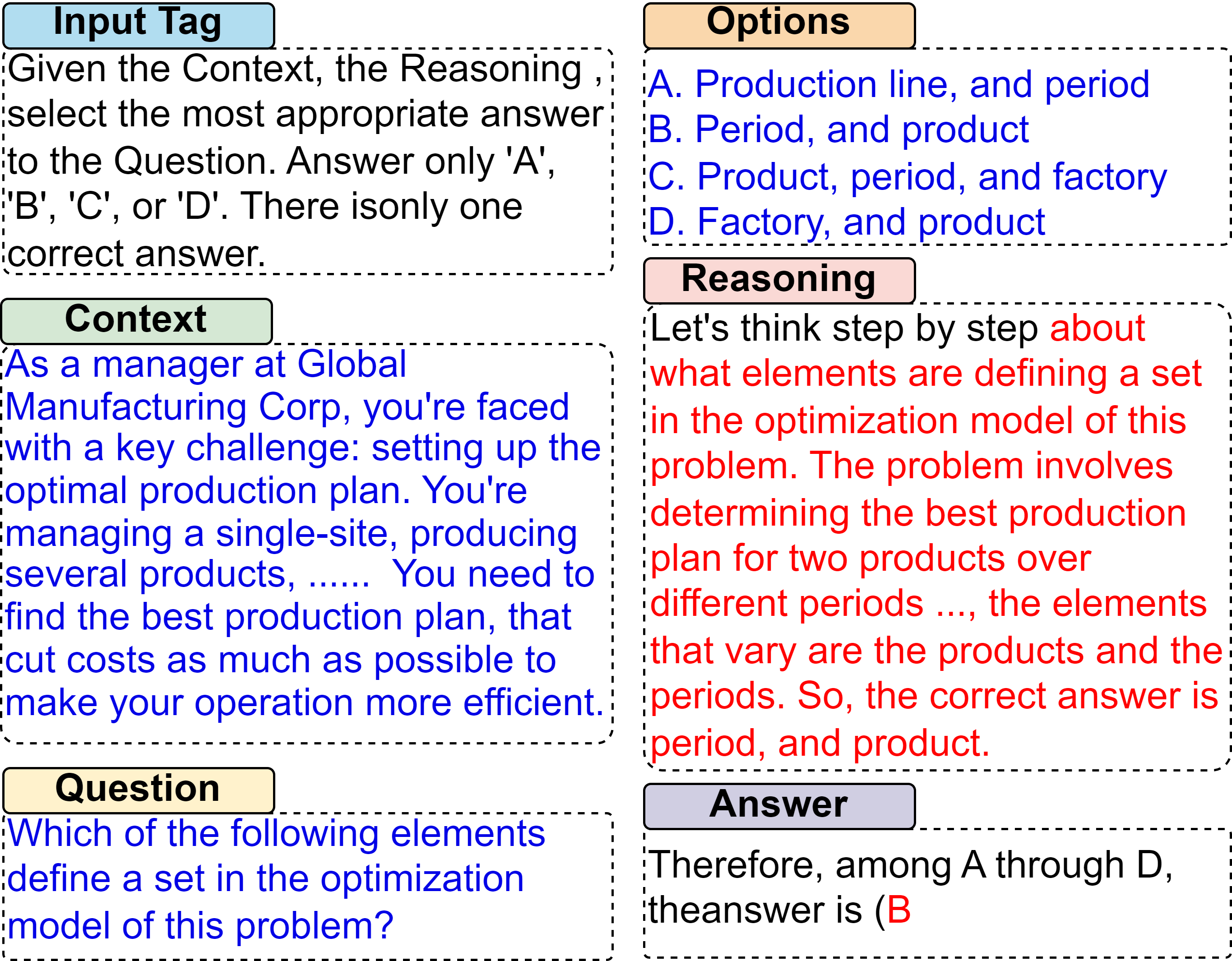}
  \caption{The different components of a prompt. The pre-defined text is in black; we provide an example (in blue), and the LLM output (in red).}
  \label{fig:prompt}
\end{figure}

An example of the different components of the prompt is shown in Figure~\ref{fig:prompt}. Standard and CoT prompting strategies are comprised of the same components shown in the figure. The only difference is that zero-shot prompts omit the explicit REASONING step, and CoT prompts would use two different INPUT TAGS between the trigger prompt and the answer-eliciting prompt. As mentioned, CoT is performed through a two-step approach. The first step extracts the REASONING component (Figure~\ref{fig:prompt}). The second step appends the REASONING component to the trigger prompt to generate the final answer.

To create the reason extraction prompt, we take the same format as the few-shot prompts, but append the TRIGGER PROMPT (e.g., ``Let's think step by step") after the list of OPTIONS.

\section{Results and Discussion}
We show the potential of ORQA as a useful benchmark for LLMs across multiple dimensions of interest. Namely, model size, model type, prompting strategy, triggering prompt, and question type. Here are some key findings:

% \textcolor{red}{\textbf{- Human experts still outperform LLMs.} As shown in Table~\ref{tab:main}, human experts achieved an accuracy of \#\#\% on a subset of ORQA. This suggests that there is still a significant gap between SOTA LLMs and human experts in specialized under-represented domains such as OR.}

\textbf{- Model size contributes to reasoning performance.} Not surprisingly, an LLM's reasoning ability is correlated to the size of an LLM. As shown in Table~\ref{tab:main}, this is true when comparing models within the same family (e.g., Llama 3.1). However, some models such as Mistral-7B and Flan-T5 perform better than Llama2-13B despite being smaller. This supports the findings from \cite{albert2023} that Mistral 7B outperforms Llama2-13B across their evaluation benchmarks potentially due to Mistral's application of sliding window attention that allows Mistral to better handle the long natural language model descriptions \citep{albert2023}. 
%Regarding FLAN-T5, it leverages CoT finetuning which has been shown to improve an LLM's reasoning abilities and may contribute to FLAN-T5 outperforming Llama2-13B (\cite{chung2024scaling}).

\begin{figure}[h!]
  \centering
  \includegraphics[width=0.9\linewidth]{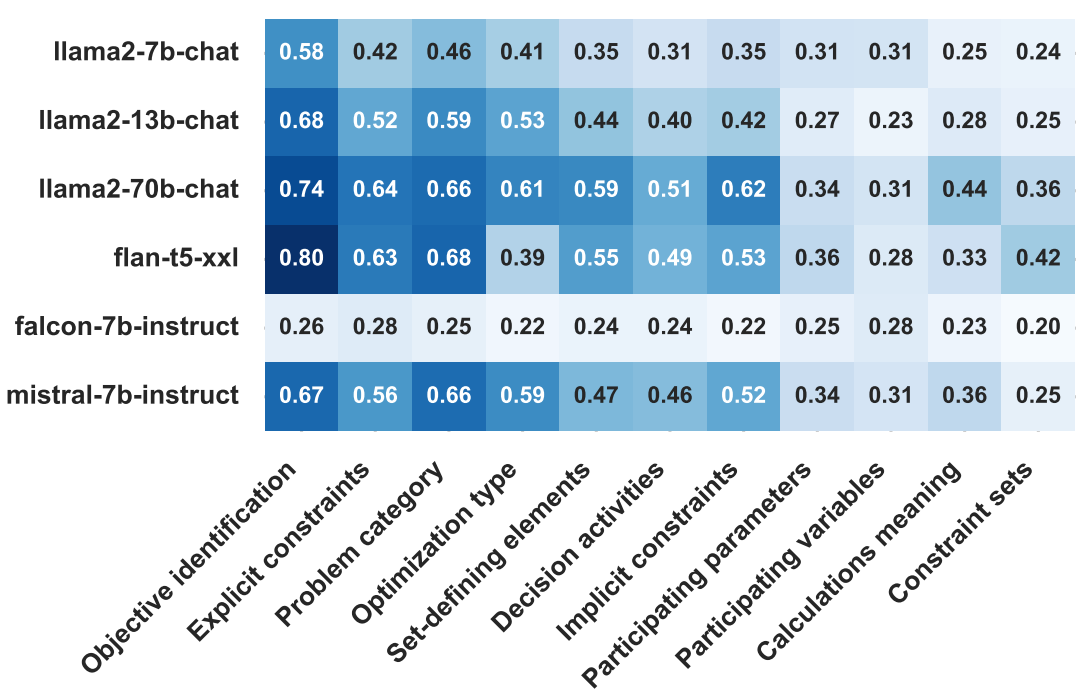}
  \caption{Heatmap of LLM performance on different question types. Performance is the average accuracy over the five prompting strategies. }
  \label{fig:heatmap}
  \includegraphics[width=0.9\linewidth]{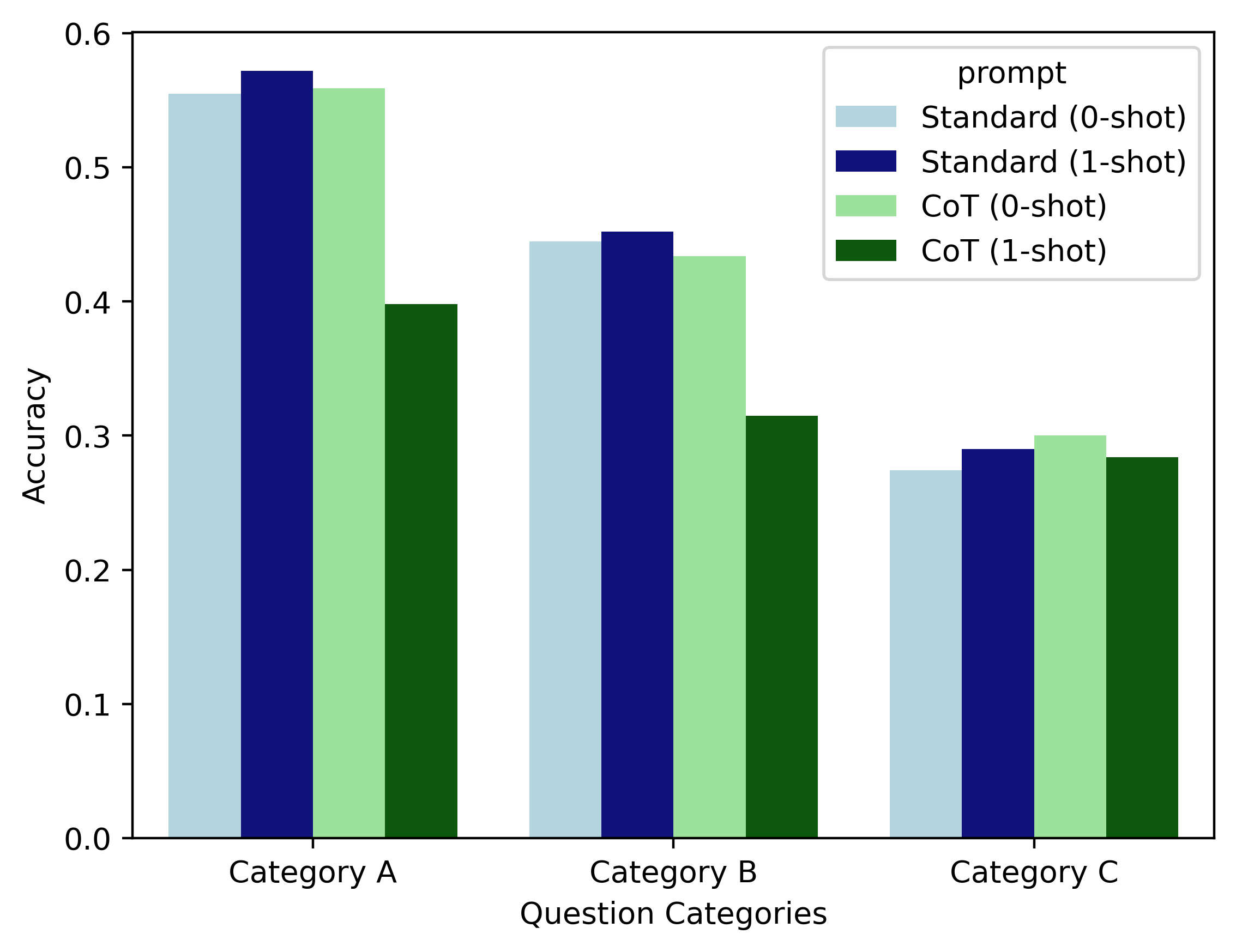}
  \caption{Performance comparison of Standard and CoT prompting. Questions of Category B \& C require more OR and/or model building knowledge.}
  \label{fig:COT_Q_categories}
\end{figure}

\begin{table*}[]
\small
    \centering
\begin{tabular}{@{}lccc@{}}
\toprule
\textbf{CoT Prompt (0-shot with Llama-3.1-70B-Instruct)}                                                                                    & \textbf{Average} & \textbf{STD} & \textbf{Best} \\ \midrule
Let's think step by step	&0.688	&0.001	&0.689 \\
Let's work by elimination	&0.648	&0.000	&0.649 \\
Let's reflect on each answer option like an operations research expert	&0.689	&0.001	&0.689\\
\begin{tabular}[c]{@{}l@{}} Let's use step by step inductive reasoning, given the mathematical \\ nature of the question\end{tabular}	&0.674	&0.003	&0.676\\
Let's think step by step like an operations research expert	&0.685	&0.000	&0.685\\ \midrule
Prompt ensembling (majority vote)  &     0.696		 &       0.008       &        0.702       \\ \bottomrule
\end{tabular}
    \caption{Impact of preceding trigger prompts on Llama-3.1-70B-Instruct using 0-shot CoT. Results are averaged over five independent runs per prompt. The bottom row shows the accuracy of ensembling all prompts.}
    \label{tab:triger}
\end{table*}

\textbf{- CoT generally drops the performance. ICL examples benefit standard but not CoT prompting.} Answering ORQA questions requires multi-step reasoning, and CoT prompting has been shown to elicit reasoning in LLMs \cite{wei2023chainofthoughtpromptingelicitsreasoning}. However, our experiments surprisingly indicate decreased performance on ORQA when using CoT (Table~\ref{tab:main} and Figure~\ref{fig:COT_Q_categories}). By investigating the generated reasoning, we found that the models often ignored instructions. For example, although the prompt specifies that only one option is correct, the models would attempt to generate reasoning that selects none or multiple options. We also report hallucinations where models would create their own options and select those. Finally, the reasonings were often incorrect, as highlighted in Figures~\ref{fig:CoT_reasoning_analysis} and \ref{fig:CoT_reasoning_analysis_validation} in Appendix~\ref{app:CoT_reasoning}. A promising direction is to explore more advanced CoT prompting techniques, including possible extensions of the faithful CoT reasoning presented by \citet{qing2023}.

\textbf{- Trigger prompts impact performance. Prompt ensembling improves CoT performance.} 
Inspired by \citet{lievin2024can}, we experimented with different trigger prompts for CoT. These experiments were conducted using the following settings: 0-shot with Llama-3.1-70B-Instruct, temperature set to 0.7, and each trigger prompt was run five times. The results are recorded in Table~\ref{tab:triger}.
Different trigger prompts can change the performance, with scores ranging from 0.648 to 0.689. %\textcolor{blue}{COMMENT: removing majority voting comment.. saying it increased by 1 percent seems insignificant and within noise..}% Applying majority voting on the results obtained from different trigger prompts provided a further $1\%$ increase in performance, reaching 0.696.

\textbf{- Fair performance on reading comprehension. Poor performance if a question requires OR knowledge and model building knowledge.} As shown in Figure~\ref{fig:heatmap}, the questions requiring only reading comprehension (left-side of the heatmap) resulted in fair performance. However, the questions that require OR and/or model building knowledge (right-side of the heatmap) were too difficult and many LLMs performed poorly. These trends can also be seen in Figure~\ref{fig:COT_Q_categories} where LLMs performed worse on the more difficult questions. To enable LLMs to deal with such complex questions is a critical venue for future work. One possibility is providing domain-specific knowledge bases or enabling API calls during reasoning, which has been shown to enhance an LLM's ability to perform knowledge-intensive reasoning tasks \cite{ReAct}. 
Another option would be to explore architecture alternatives to transformers such as StripedHyena \citep{poli2023stripedhyena}, that could better deal with the concerning findings of \citet{dziri2024faith} which suggest transformer-based LLMs reduce compositional reasoning into linearized subgraph matching. 
In essence, these findings seem to indicate that transformer-based LLMs are inherently limited in their ability to perform complex multi-step reasoning.
Along these lines, ORQA could represent a useful benchmark for testing LLMs based on those novel architectures.

\textbf{- In-context learning can reduce the number of reasoning steps and reasoning errors.}
We analyzed reasoning steps generated by Llama2-13b-chat model in 0-shot and 1-shot CoT settings on our validation set. OR experts manually evaluated a total of 90 instances (45 per setting) and classified the reasoning into four categories: correct reasoning, incorrect logic, insufficient knowledge, and incorrect reading comprehension. Representative examples are provided in Appendix~\ref{app:CoT_reasoning} (Figure~\ref{fig:CoT_reasoning_analysis_validation}). As shown in Table~\ref{tab:reasoning_stats}, the number of reasoning steps in the 0-shot setting is significantly higher than in the 1-shot setting. This increases the likelihood of errors in intermediate steps, explaining the lower accuracy in 0-shot reasoning. Interestingly, in 20\% of 0-shot tests, the model arrived at the correct answer despite some reasoning errors. Lastly, correct reasoning strongly correlates with finding the correct answer.
Please refer to Appendix~\ref{app:error_type} for our analysis on the relations between reasoning error types and question categories.
\begin{table}[h]
\small
\centering
\begin{tabular}{lcc}
\toprule
\textbf{Metric} & \textbf{0-shot} & \textbf{1-shot} \\
\midrule
Instances with correct answer & 35.6\% & 33.3\% \\
\begin{tabular}[c]{@{}l@{}} Instances where all are reasoning \\ \hspace{10pt} steps are correct\end{tabular}& 15.6\% & 31.1\% \\
Incorrect reasoning, correct answer & 20.0\% & 6.7\% \\
Incorrect answer, correct reasoning & 0.0\% & 4.4\% \\
Avg. number of steps per instance & 7.93 & 4.53 \\
Avg. accuracy of steps per instance & 0.682 & 0.611 \\
\bottomrule
\end{tabular}
\caption{Statistics on generated reasoning steps and answers using Llama2-13b-Chat on the validation set with CoT.}
\label{tab:reasoning_stats}
\end{table}

\textbf{- Length of ICL examples has a greater influence on accuracy than the similarity of question types.} We explore the impact of ICL example selection on ORQA by running four experiments on the Llama2-13B-Chat with 1-shot CoT: (1) random ICL selection, (2) same question type, (3) similar prompt length, (4) same question type and similar prompt length. Table~\ref{tab:ICL_examples} presents the results. We found that while both the length of ICL examples and the similarity of question types improved the performance, an example of similar length had more impact. 

\begin{table}[h]
\centering
\small
\begin{tabular}{lc}
\hline
\textbf{Approach} & \textbf{Accuracy} \\
\hline
Random selection & 0.300 \\
Same question type & 0.313 \\
Similar length & 0.353 \\
Similar length \& same question type & 0.362 \\
\hline
\end{tabular}
\caption{Comparison of different ICL example selection approaches.}
\label{tab:ICL_examples}
\end{table}

\section{Conclusion and Future Works}
LLMs have received recognition and popularity through interfaces like ChatGPT. However, as they are being used more in high-stakes fields such as medicine and law, we must understand the reasoning behind these LLM responses. Thus, we developed the Operations Research QA (ORQA) benchmark and utilized it to evaluate the generalization and reasoning abilities of some popular pretrained LLMs within a novel and technically complex domain. We explored different prompting strategies to evaluate perspectives of reasoning using this multi-choice QA benchmark dataset. Our results show that there is still considerable room for improvement across different LLMs with Llama3.1-405B-Instruct achieving the highest accuracy of 0.772 and human expert accuracy (on a subset) of 0.93. % With these modest results, we encourage LLM developers and researchers to leverage ORQA when evaluating the ability of an LLM to generalize to a new specialized technical domain.

We acknowledge that structured reasoning metrics, such as those in ROSCOE \cite{golovneva2023roscoesuitemetricsscoring}, reasoning graph accuracy and similarity in STREET \cite{ribeiro2023streetmultitaskstructuredreasoning}, and entailment tree comparisons \cite{dalvi-etal-2021-explaining}, are promising for automating the evaluation of reasoning steps in ORQA. By making ORQA publicly available, we look forward to seeing it used as a benchmark for LLMs, especially for models trained specifically on the task of QA or OR and related fields. This would provide additional clarity to the difficulty of this benchmark and how general language models perform against domain or task-specific models. We also encourage experts to expand this dataset by introducing not only more OR multi-choice QA problems, but also additional multi-choice QA problems from other technical fields similarly requiring expert-level knowledge and reasoning skills with limited publicly available text corpora.

{\small
\bibliography{aaai25}
}

\newpage
\appendix

\section{Optimization problem components and their relationships}\label{app:components}

\begin{figure}[H]
  \centering
  \includegraphics[width=0.8\linewidth]{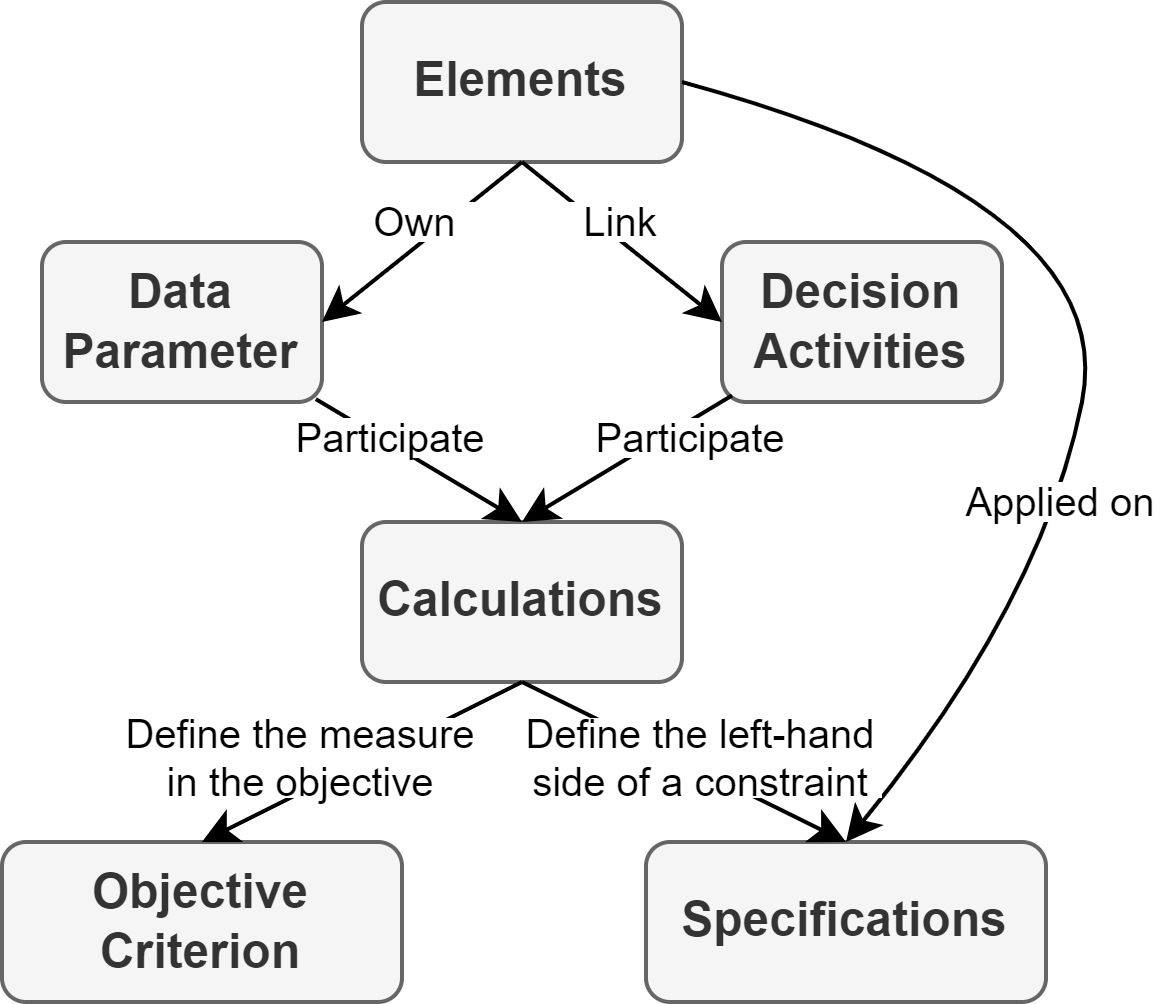}
\caption{The components of an optimization problem and their relationships}
\label{Fig:components}
\end{figure}

\section{F1 score results}\label{app:F1}
Table~\ref{tab:F1} reports the F1 scores for several models. We did not observe a meaningful difference between accuracy and F1, indicating that there is no imbalance in the distribution of target answers across A, B, C, and D.

\begin{table}[h]
\small
\centering
\begin{tabular}{@{}l|l|llc|ll@{}}
\toprule
\multicolumn{1}{c|}{\multirow{2}{*}{\textbf{Model}}} & \multirow{2}{*}{\textbf{Metric}} & \multicolumn{3}{c|}{\textbf{Standard (Acc)}} & \multicolumn{2}{c}{\textbf{CoT (Acc)}} \\ \cmidrule(l){3-7} 
\multicolumn{1}{c|}{} &  & \multicolumn{1}{c}{\textbf{\begin{tabular}[c]{@{}c@{}}0-\\ shot\end{tabular}}} & \multicolumn{1}{c}{\textbf{\begin{tabular}[c]{@{}c@{}}1-\\ shot\end{tabular}}} & \textbf{\begin{tabular}[c]{@{}c@{}}3-\\ shot\end{tabular}} & \multicolumn{1}{c}{\textbf{\begin{tabular}[c]{@{}c@{}}0-\\ shot\end{tabular}}} & \multicolumn{1}{c}{\textbf{\begin{tabular}[c]{@{}c@{}}1-\\ shot\end{tabular}}} \\ \midrule
\multirow{2}{*}{Llama3.1-8B-I} & Acc & 0.588 & 0.615 & 0.619 & 0.563 & 0.324 \\
 & F1 & 0.588 & 0.615 & 0.618 & 0.563 & 0.321 \\ \midrule
\multirow{2}{*}{Llama3.1-70B-I} & Acc & 0.702 & 0.721 & 0.735 & 0.689 & 0.292 \\
 & F1 & 0.702 & 0.721 & 0.735 & 0.689 & 0.287 \\ \midrule
\multirow{2}{*}{Llama3.1-405B-I} & Acc & 0.723 & 0.753 & 0.772 & 0.695 & 0.360 \\
 & F1 & 0.723 & 0.753 & 0.772 & 0.695 & 0.355 \\ \midrule
\multirow{2}{*}{Llama3-8B-I} & Acc & 0.535 & 0.573 & 0.592 & 0.530 & 0.364 \\
 & F1 & 0.533 & 0.574 & 0.591 & 0.533 & 0.361 \\ \midrule
\multirow{2}{*}{Llama3-70B-I} & Acc & 0.676 & 0.716 & 0.710 & 0.671 & 0.448 \\
 & F1 & 0.676 & 0.716 & 0.711 & 0.672 & 0.448 \\ \midrule
\multirow{2}{*}{Mistral-7B-I-v0.3} & Acc & \multicolumn{1}{c}{0.523} & \multicolumn{1}{c}{0.555} & 0.555 & \multicolumn{1}{c}{0.539} & \multicolumn{1}{c}{0.543} \\
 & F1 & \multicolumn{1}{c}{0.519} & \multicolumn{1}{c}{0.552} & 0.554 & \multicolumn{1}{c}{0.538} & \multicolumn{1}{c}{0.543} \\ \midrule
\multirow{2}{*}{Mixtral-8x7B-I-v0.1} & Acc & \multicolumn{1}{c}{0.589} & \multicolumn{1}{c}{0.606} & 0.612 & \multicolumn{1}{c}{0.565} & \multicolumn{1}{c}{0.565} \\
 & F1 & \multicolumn{1}{c}{0.588} & \multicolumn{1}{c}{0.605} & 0.612 & \multicolumn{1}{c}{0.564} & \multicolumn{1}{c}{0.502} \\ \bottomrule
\end{tabular}
\caption{Accuracy and Macro-averaged F1 score for Llama, Mistral and Mixtral models across different prompting
strategies}
\label{tab:F1}
\end{table}

\section{Reasoning Error Types and Question Category}\label{app:error_type}
As expected category A, has the largest number of correct steps (77.2\% of cases). Category B showed a lower percentage of correct steps at 69.0\%, with higher occurrences of incorrect logic (13.4\%) and incorrect reading comprehension (7.0\%). Category C had 74.0\% correct steps, with similar incorrect logic and incorrect reading comprehension steps to Category B.

\begin{table}[H]
\small
\centering
\begin{tabular}{@{}lcccc@{}}
\toprule
\textbf{\begin{tabular}[c]{@{}l@{}}Question \\ Category\end{tabular}} & \textbf{\begin{tabular}[c]{@{}c@{}}Correct \\ Steps\end{tabular}} & \textbf{\begin{tabular}[c]{@{}c@{}}Incorrect\\  Logic\end{tabular}} & \textbf{\begin{tabular}[c]{@{}c@{}}Insufficient\\  Knowledge\end{tabular}} & \textbf{\begin{tabular}[c]{@{}c@{}}Incorrect \\ Reading \\ Comprehension\end{tabular}} \\ \midrule
Category A & 0.772 & 0.087 & 0.102 & 0.039 \\
Category B & 0.690 & 0.134 & 0.106 & 0.070 \\
Category C & 0.740 & 0.134 & 0.063 & 0.063 \\ \bottomrule
\end{tabular}
\caption{Reasoning Error Types by Question Category}
\label{tab:reasoning_error_types}
\end{table}

\section{Question types and categories}\label{app:categories}

Table~\ref{tab:Q_category} introduces the three question categories, the skills required to answer questions in each category, related question types, as well as examples and descriptions of the questions. 

\begin{table*}[h!]
\small
\centering
\begin{tabular}{@{}cccll@{}}
\toprule
\textbf{Category} & \textbf{Critical skills} & \textbf{Question name} & \textbf{Description} & \textbf{Question example} \\ \midrule
\multirow{3}{*}{\begin{tabular}[c]{@{}c@{}}Category A:\\ \\ Understanding \\ the high-level \\ problem \\ specifications\end{tabular}} &  & \begin{tabular}[c]{@{}c@{}}Q1: \\ Objective identification\end{tabular} & \begin{tabular}[c]{@{}l@{}}Identify the objective of \\ the optimization model\end{tabular} & \begin{tabular}[c]{@{}l@{}}Which of the following\\ define the objective criterion\\ of the problem?\end{tabular} \\ \cmidrule(l){3-5} 
 & \begin{tabular}[c]{@{}c@{}}Reading \\ comprehension\end{tabular} & \begin{tabular}[c]{@{}c@{}}Q2: \\ Explicit constraints\end{tabular} & \begin{tabular}[c]{@{}l@{}}Identify the constraints \\ of the optimization model \\ that are explicitly stated \\ in the context\end{tabular} & \begin{tabular}[c]{@{}l@{}}Which of the following options \\ define a single assignment \\ constraint that is required \\ for this problem?\end{tabular} \\ \cmidrule(l){3-5} 
 &  & \begin{tabular}[c]{@{}c@{}}Q3: \\ Problem category\end{tabular} & \begin{tabular}[c]{@{}l@{}}Identify the most related \\ basic optimization problem \\ to the context\end{tabular} & \begin{tabular}[c]{@{}l@{}}Under which category \\ does the given optimization \\ problem fall into?\end{tabular} \\ \midrule
\multirow{4}{*}{\begin{tabular}[c]{@{}c@{}}Category B:\\ \\ Identifying \\ entities of the \\ corresponding \\ optimization \\ model\end{tabular}} &  & \begin{tabular}[c]{@{}c@{}}Q4: \\ Optimization type\end{tabular} & \begin{tabular}[c]{@{}l@{}}Identify the type of \\ optimization model, \\ such as linear, \\ non-linear, or \\ integer programming\end{tabular} & \begin{tabular}[c]{@{}l@{}}What is the type of \\ optimization model \\ related to this problem?\end{tabular} \\ \cmidrule(l){3-5} 
 & \begin{tabular}[c]{@{}c@{}}Reading \\ comprehension\end{tabular} & \begin{tabular}[c]{@{}c@{}}Q5: \\ Set-defining elements\end{tabular} & \begin{tabular}[c]{@{}l@{}}Identify the set of \\ the optimization model\end{tabular} & \begin{tabular}[c]{@{}l@{}}Which of the following defines \\ a set in the optimization model\\  of this problem?\end{tabular} \\ \cmidrule(l){3-5} 
 & OR knowledge & \begin{tabular}[c]{@{}c@{}}Q6: \\ Decision activities\end{tabular} & \begin{tabular}[c]{@{}l@{}}Identify the main \\ variables of the \\ optimization model\end{tabular} & \begin{tabular}[c]{@{}l@{}}What are the decision\\ activities of the \\ optimization problem?\end{tabular} \\ \cmidrule(l){3-5} 
 &  & \begin{tabular}[c]{@{}c@{}}Q7: \\ Implicit constraints\end{tabular} & \begin{tabular}[c]{@{}l@{}}Find constraints of the \\ optimization model that \\ are not explicitly mentioned \\ in the context but \\ needed for the model\end{tabular} & \begin{tabular}[c]{@{}l@{}}Which of the following \\ constraints are required \\ to properly formulate \\ this optimization problem?\end{tabular} \\ \midrule
\multirow{4}{*}{\begin{tabular}[c]{@{}c@{}}Category C:\\ \\  Identifying \\ relationships \\ between entities\end{tabular}} & \begin{tabular}[c]{@{}c@{}}Reading \\ comprehension\end{tabular} & \begin{tabular}[c]{@{}c@{}}Q8: \\ Participating parameters\end{tabular} & \begin{tabular}[c]{@{}l@{}}Identify the parameters\\ involved in the expression \\ of the objective of \\ the optimization model\end{tabular} & \begin{tabular}[c]{@{}l@{}}Which data parameters are \\ participating in the objective \\ criterion for this problem?\end{tabular} \\ \cmidrule(l){3-5} 
 & OR knowledge & \begin{tabular}[c]{@{}c@{}}Q9: \\ Participating variables\end{tabular} & \begin{tabular}[c]{@{}l@{}}Identify the variables \\ involved in the expression \\ of the objective of the \\ optimization model\end{tabular} & \begin{tabular}[c]{@{}l@{}}Which of the following are \\ participating decision \\ activities in the objective\\ criterion for this problem?\end{tabular} \\ \cmidrule(l){3-5} 
 & Model building & \begin{tabular}[c]{@{}c@{}}Q10: \\ Calculations meaning\end{tabular} & \begin{tabular}[c]{@{}l@{}}Explain the meaning of the\\ calculations which are \\ intermediate expressions in \\ the optimization model\end{tabular} & \begin{tabular}[c]{@{}l@{}}In this problem, there is a \\ production capacity; \\ what is the meaning of\\  its left-hand side?\end{tabular} \\ \cmidrule(l){3-5} 
 &  & \begin{tabular}[c]{@{}c@{}}Q11: \\ Constraint sets\end{tabular} & \begin{tabular}[c]{@{}l@{}}Identify the set that a given\\ constraint is applied on\end{tabular} & \begin{tabular}[c]{@{}l@{}}In this problem, on which of \\ the following system \\ elements are inventory \\ balance constraints applied?\end{tabular} \\ \bottomrule
\end{tabular}
\caption{Question types, categories, and required skills.}
\label{tab:Q_category}
\end{table*}

\section{Illustrative example for dataset generation steps}\label{app:dataset}

\begin{figure*}[h!]
  \centering
  \includegraphics[width=\linewidth]{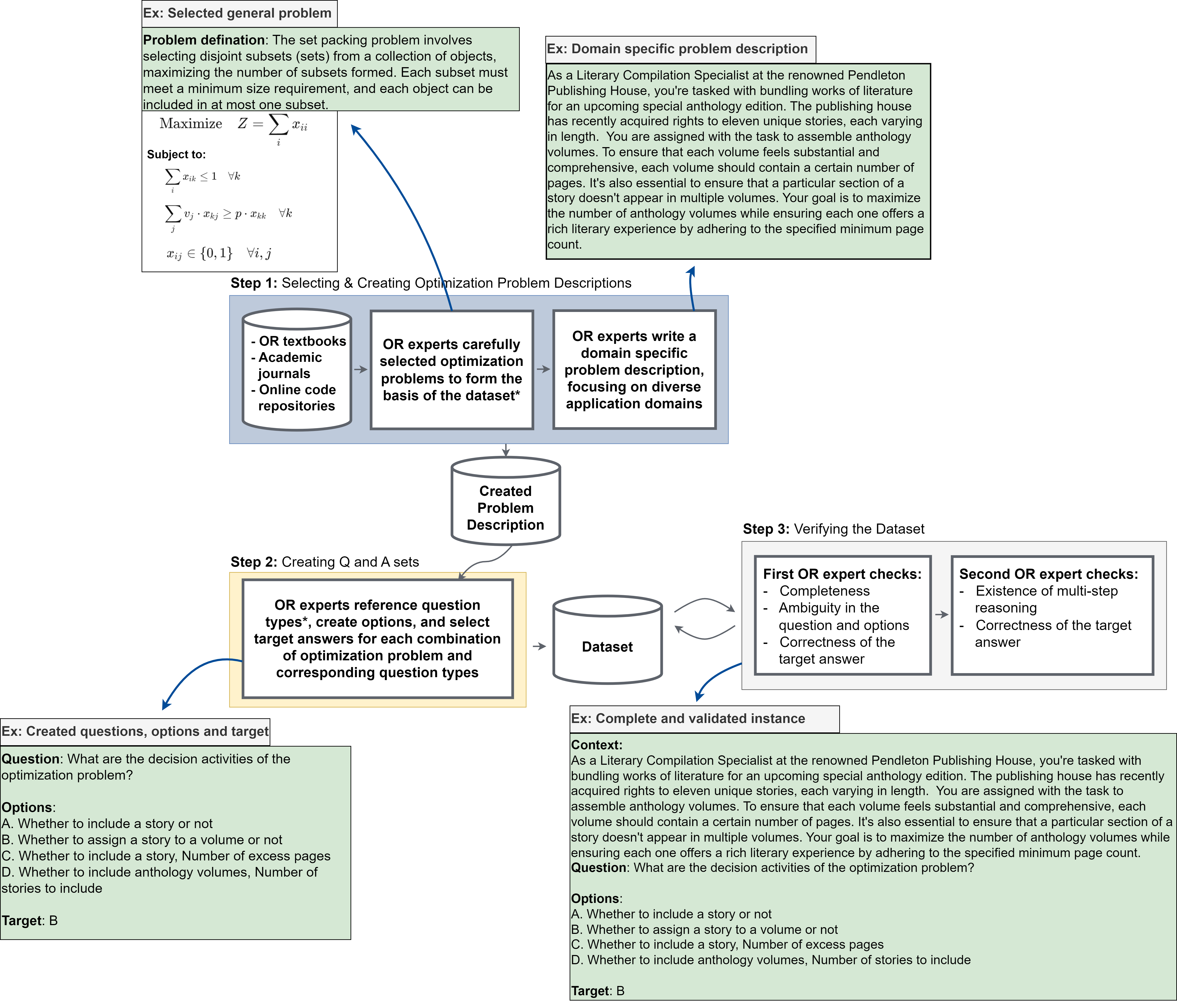}
  \caption{Green boxes show how a dataset instance is generated in our three-step dataset generation process.}
  \label{fig:dataCreation_exp}
\end{figure*}

Figure~\ref{fig:dataCreation_exp} clarifies the dataset creation steps with an example.

\section{Illustrative examples of CoT reasoning annotation} \label{app:CoT_reasoning}

\begin{figure*}[h!]
  \centering
  \includegraphics[width=\linewidth]{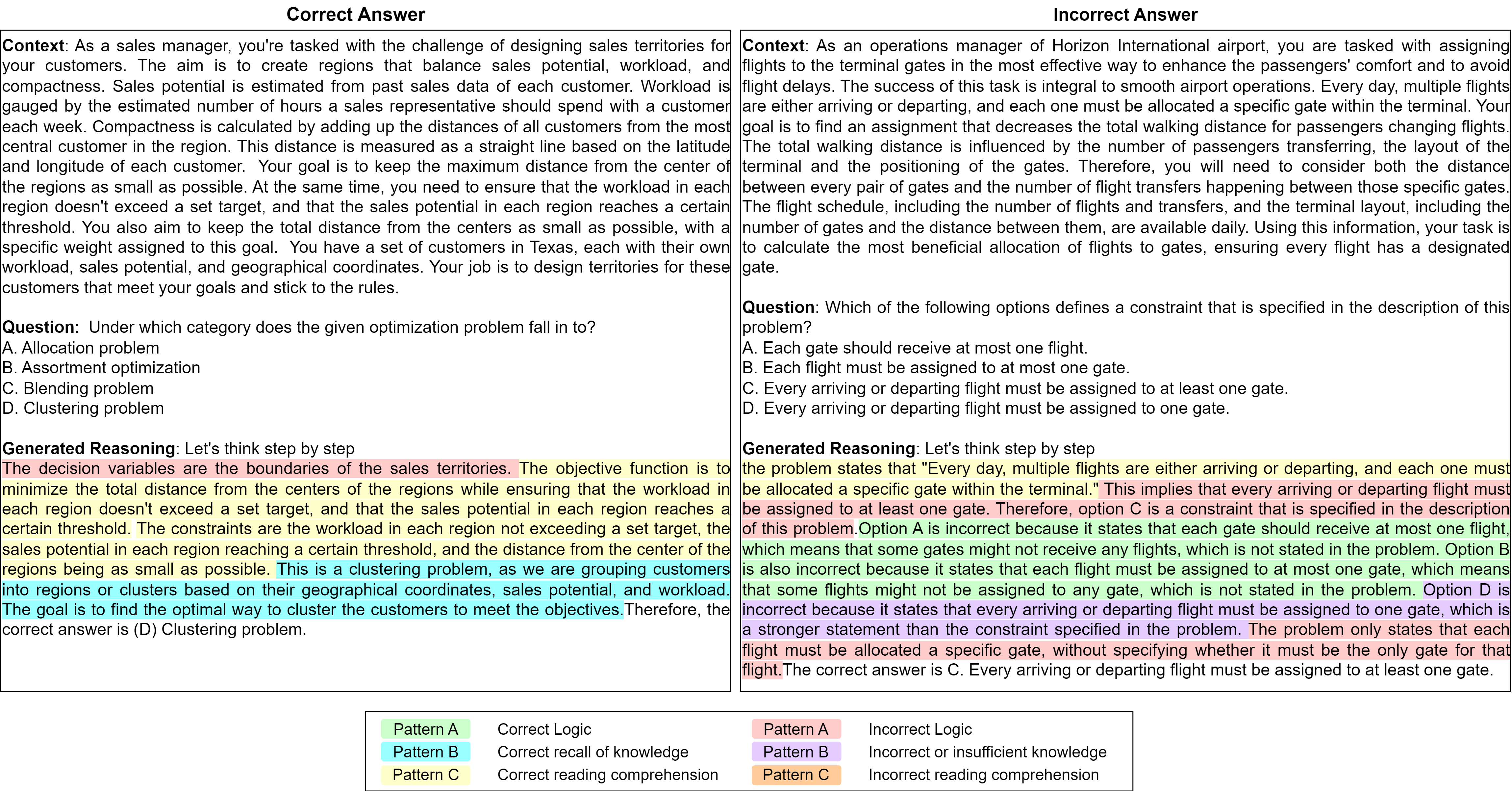}
  \caption{Annotation of CoT reasoning for two illustrative examples. Right: Although there is one incorrect reasoning step, the model answer is correct (option D). Left: The model made multiple mistakes in the reasoning steps, resulting in a wrong answer. The correct option is A. The reasoning steps are generated using Llama2-13B-Chat in a 0-shot CoT setting.}
  \label{fig:CoT_reasoning_analysis}
\end{figure*}

\begin{figure*}[h!]
  \centering
  \includegraphics[width=\linewidth]{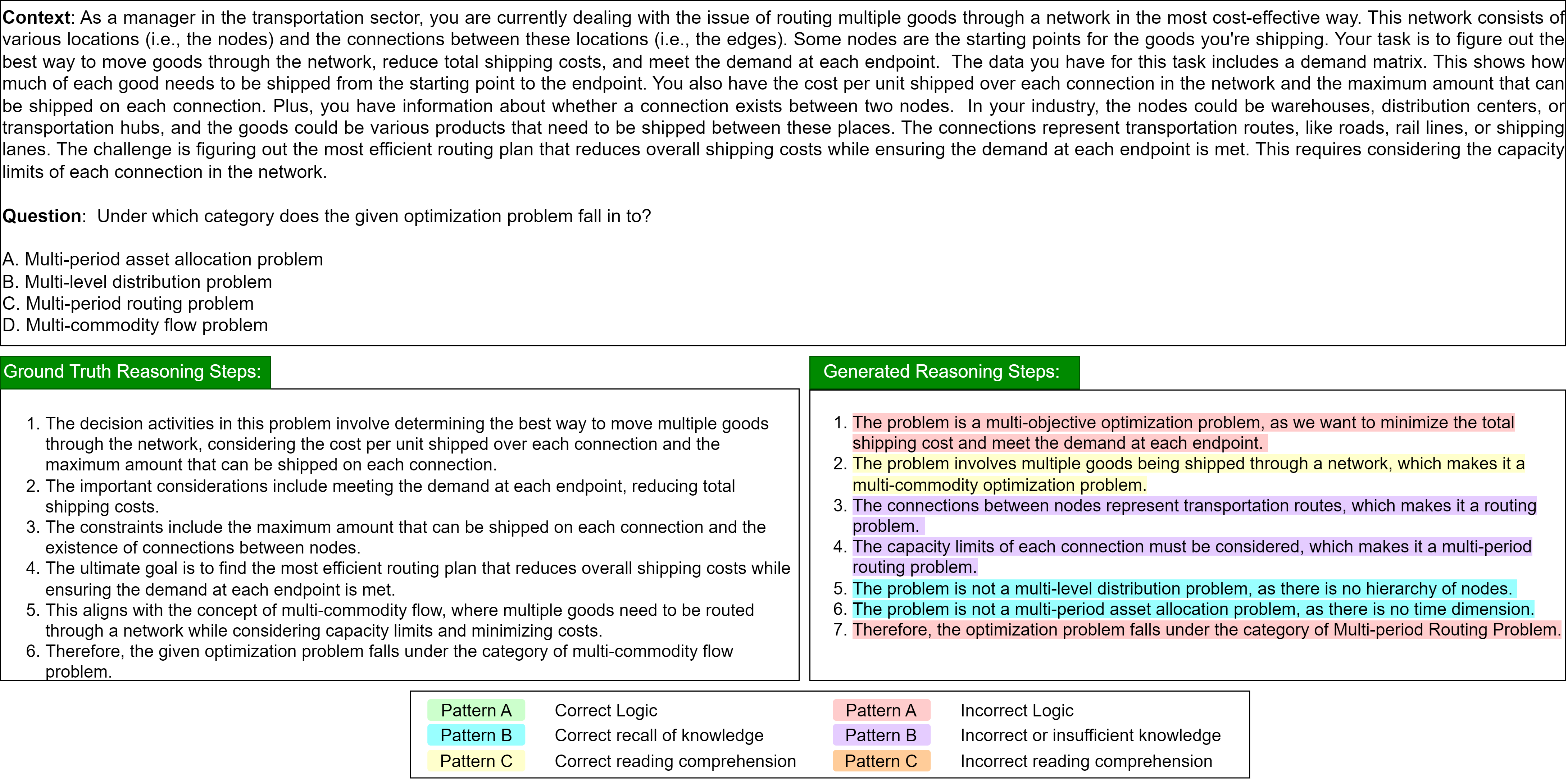}
  \caption{Annotation of CoT reasoning for an illustrative example from ORQA validation set, where we have ground truth reasoning. The reasoning steps are generated using Llama2-13B-Chat in a 0-shot CoT setting.}
  \label{fig:CoT_reasoning_analysis_validation}
\end{figure*}

Figures~\ref{fig:CoT_reasoning_analysis} and \ref{fig:CoT_reasoning_analysis_validation} illustrate the reasoning steps generated by a model (e.i. Llama2-13B-Chat) and the common errors made in those reasoning steps.

\end{document}